\documentclass[a4paper]{article}
\usepackage{amsmath,graphicx}
\usepackage{bm}
\usepackage{cite}

\newcommand{\argmin}{\mathop{\rm argmin}\limits}
\usepackage{multirow}

\usepackage{INTERSPEECH2021}

\title{Unified Autoregressive Modeling for Joint End-to-End Multi-Talker \\Overlapped Speech Recognition and Speaker Attribute Estimation} 
\name{Ryo Masumura$^\dagger$, Daiki Okamura$^{\dagger\ddagger}$, Naoki Makishima$^\dagger$, Mana Ihori$^\dagger$,\\
  Akihiko Takashima$^\dagger$, Tomohiro Tanaka$^\dagger$, Shota Orihashi$^\dagger$}
\address{$^\dagger$ NTT Media Intelligence Laboratories, NTT Corporation, Japan\\
$^\ddagger$ Nagaoka University of Technology, Japan}
\email{ryou.masumura.ba@hco.ntt.co.jp}

\begin{document}

\maketitle
\begin{abstract}
  In this paper, we present a novel modeling method for single-channel multi-talker overlapped automatic speech recognition (ASR) systems. Fully neural network based end-to-end models have dramatically improved the performance of multi-taker overlapped ASR tasks. One promising approach for end-to-end modeling is autoregressive modeling with serialized output training in which transcriptions of multiple speakers are recursively generated one after another. This enables us to naturally capture relationships between speakers. However, the conventional modeling method cannot explicitly take into account the speaker attributes of individual utterances such as gender and age information. In fact, the performance deteriorates when each speaker is the same gender or is close in age. To address this problem, we propose unified autoregressive modeling for joint end-to-end multi-talker overlapped ASR and speaker attribute estimation. Our key idea is to handle gender and age estimation tasks within the unified autoregressive modeling. In the proposed method, transformer-based autoregressive model recursively generates not only textual tokens but also attribute tokens of each speaker. This enables us to effectively utilize speaker attributes for improving multi-talker overlapped ASR. Experiments on Japanese multi-talker overlapped ASR tasks demonstrate the effectiveness of the proposed method.
\end{abstract}
\noindent\textbf{Index Terms}: multi-talker ASR, speaker attribute estimation, unified autoregressive modeling 

\section{Introduction}
Typical automatic speech recognition (ASR) systems are single-talker systems that convert a single speaker's monaural speech signal into a single utterance transcription. On the other hand, in natural conversations and meetings, such single-talker systems limit their applicability because multiple utterances are often overlapped. Therefore, recognizing multi-talker overlapped monaural speech signals has been the focus of much attention recently. So far, multi-talker overlapped ASR problems are designed as a cascade system of speech separation \cite{hershey_icassp2016,yu_icassp2017} and single-talker ASR. However, the cascade system is not necessarily optimal for multi-talker overlapped ASR tasks because speech separation modules are often optimized by a signal-level criterion.

With the progress of deep learning technology, end-to-end models have become a common alternative to traditional hybrid models in single-talker ASR tasks \cite{bahdanau_icassp2015,lu_interspeech2015,masumura_icassp2019,dong_icassp2018,masumura_icassp2021}. Similarly, end-to-end models have dramatically improved the performance of multi-taker overlapped ASR tasks compared with the cascade system. Initial end-to-end multi-talker overlapped ASR systems use permutation invariant training (PIT), which can solve the label-permutation problem by considering all possible permutations of speakers \cite{yu_interspeech2017,chang_icassp2019,seki_acl2018,settle_icassp2018,chang_icassp2020,sklyar_arxiv2020}. In fact, the PIT-based model must have multiple output layers corresponding to different speakers. Thus, one weakness of PIT-based modeling is that it cannot handle the dependency among utterances of multiple speakers because the output layers are independent from each other. In contrast, a recent hopeful approach for the end-to-end modeling is autoregressive modeling with serialized output training (SOT) in which transcriptions of multiple speakers are recursively generated one after another from single output layer \cite{kanda_interspeech2020a}. The main advantage of the modeling is that we can naturally model the dependency among the outputs for multiple speakers, which could help avoid duplicate hypotheses from being generated.

The autoregressive modeling with SOT has performed impressively. However, it cannot explicitly take into account the speaker attributes of individual utterances such as gender and age information. In fact, the conventional method \cite{kanda_interspeech2020a} only captures textual information of individual utterances. Therefore, the performance deteriorates when each speaker is the same gender or is close in age. To mitigate this problem, we aim to use not only the textual information but also speaker attributes as contexts in a single-channel multi-talker overlapped ASR system. Only few studies have tried to capture speaker information by jointly modeling with speaker identification \cite{shafey_interspeech2019,mao_interspeech2020,kanda_interspeech2020b}. Unfortunately, joint modeling with speaker identification is not suitable for handling multiple arbitrary speakers because these studies are specialized for handling known speakers or known speaker roles. In other words, they have limitations that speaker labels cannot be estimated for unknown speakers.  

In this paper, we propose a novel method that jointly models end-to-end multi-talker overlapped ASR and speaker attribute estimation. Our key idea is to handle gender and age estimation tasks within a unified autoregressive modeling. Unlike handling the speaker labels, the speaker attributes can be estimated for arbitrary speakers. In the proposed method, the transformer-based autoregressive model recursively generates not only textual tokens but also attribute tokens of each speaker. This helps us to effectively utilize speaker attributes for capturing the dependency among utterances of multiple speakers. In fact, a similar idea has been proposed for multilingual multi-speaker overlapped ASR \cite{seki_interspeech2019}, where PIT-based multi-talker overlapped ASR is combined with a language identification task. But, it cannot handle the dependency among utterances of multiple speakers. To the best of our knowledge, this paper is the first to jointly model multi-talker overlapped ASR and auxiliary tasks using unified autoregressive modeling. In experiments on Japanese multi-talker overlapped ASR tasks, we show that estimating speaker attribute information improves multi-talker overlapped ASR performance.

\section{Related Work}

\noindent {\bf Speaker attribute estimation:}
In speech fields, various methods that estimate speaker attributes such as gender, age, and height have been studied \cite{li_2013,grzybowska_2016,ghahremani_2018,markitantov_2019,kalluri_2020,sarma_2020}. In the last decade, fully neural network based methods have been examined to precisely capture input speech contexts \cite{ghahremani_2018,markitantov_2019,kalluri_2020,sarma_2020}. In fact, multiple-speaker attributes are often jointly estimated via multi-task learning \cite{markitantov_2019,sarma_2020}. In this paper, we jointly estimate gender and age information with textual information in multi-talker overlapped ASR. This paper is the first to estimate multiple-speaker attributes from the overlapped speech and to utilize the estimated speaker attributes to improve multi-talker overlapped ASR.

\smallskip
\noindent {\bf Token-augmented speech recognition:}
In several studies, special tokens are augmented for extending end-to-end ASR modeling. To jointly learn multilingual end-to-end ASR problems, language identification and end-to-end ASR are jointly modeled using language tokens \cite{watanabe_asru2017,seki_interspeech2019}. In addition, to consider various speech signals, social signal tokens are jointly estimated with textual tokens in end-to-end ASR \cite{inaguma_icassp2018}. Furthermore, end-to-end ASR, audio tagging, and acoustic event detection are jointly modeled using task tokens and event tokens \cite{moritz_interspeech2020}. In end-to-end multi-talker overlapped ASR, a separator token that represents the speaker change is augmented \cite{kanda_interspeech2020a}. In this paper, in addition to the separator token, we augment gender tokens and age tokens to jointly model speaker attribute estimation and multi-talker overlapped ASR.

\section{Conventional Methods}
This section details single-channel single-talker ASR and single-channel multi-talker overlapped ASR with autoregressive modeling.
Note that fully neural network based autoregressive modeling is omitted in this section (see section 4.2).

\subsection{Single-talker ASR with autoregressive modeling}
The single-talker ASR with autoregressive modeling predicts the generation probability of textual tokens $\bm{W}=\{w_1,\cdots,w_N\}$ given monaural speech $\bm{X}=\{\bm{x}_1,\cdots,\bm{x}_M\}$, where $w_n \in {\cal V}$ is the $n$-th token in the textual tokens and $\bm{x}_m$ is the $m$-th acoustic feature in the speech. $N$ is the number of tokens in the output, $M$ is the number of acoustic features in the speech, and ${\cal V}$ is the vocabulary set. In the autoregressive modeling, the generation probability of $\bm{W}$ is defined as
\begin{equation}
  P(\bm{W}|\bm{X}; \bm{\Theta}_{\tt st}) = \prod_{n=1}^N P(w_n|w_{1:n-1}, \bm{X}; \bm{\Theta}_{\tt st}) ,
\end{equation}
where $\bm{\Theta}_{\tt st}$ represents the trainable model parameter sets and $w_{1:n-1}=\{w_1,\cdots,w_{n-1}\}$. 
The model parameter set can be optimized by
\begin{equation}
  \hat{\bm{\Theta}}_{\tt st}  = \argmin_{\bm{\Theta}_{\tt st}} - \sum_{(\bm{X},\bm{W}) \in {\cal D}_{\tt st}} \log P(\bm{W}|\bm{X}; \bm{\Theta}_{\tt st}) , 
\end{equation}
where ${\cal D}_{\tt st}$ represents the single-talker speech dataset.

\subsection{Multi-talker overlapped ASR with autoregressive modeling}
The multi-talker overlapped ASR with autoregressive modeling predicts the generation probability of multiple utterance-level textual tokens $\bm{W}^{1:T} = \{\bm{W}^1,\cdots,\bm{W}^T\}$ for each speaker given overlapped monaural speech $\bm{X}$, where $\bm{W}^t=\{w_1^t,\cdots,w_{N^t}^t\}$ is the $t$-th speaker's textual tokens, $N^t$ is the number of tokens in the $t$-th speaker's textual tokens and $T$ is the number of speakers in the overlapped monaural speech. There are multiple permutations in the order of the multiple outputs $\bm{W}^{1:T}$, so we sort these multiple outputs by their start times, which is called first-in, first-out \cite{tripathi_2020,kanda_interspeech2020a}. In this case, the generation probability of $\bm{W}^{1:T}$ is defined as
\begin{equation}
  P(\bm{W}^{1:T} |\bm{X}; \bm{\Theta}_{\tt mt}) = \prod_{t=1}^T P(\bm{W}^t|\bm{W}^{1:t-1}, \bm{X};  \bm{\Theta}_{\tt mt}) ,
\end{equation}
where $\bm{\Theta}_{\tt mt}$ represents the trainable model parameter sets. In this way, textual tokens for individual speakers are recursively generated one after another in an autoregressive manner. The main advantage of the modeling is that we can naturally model the dependency among the individual output textual tokens. In autoregressive modeling with SOT, to recognize multiple utterances efficiently, multiple textual tokens are serialized into a single token sequence. Thus, the generation probability of $\bm{W}^{1:T}$ is redefined as
\begin{equation}
  \begin{split}
  P(\bm{W}^{1:T} |\bm{X}; \bm{\Theta}_{\tt mt}) & =  P(\bm{S} |\bm{X}; \bm{\Theta}_{\tt mt})  \\
  & = \prod_{l=1}^{|\bm{S}|} P(s_l|s_{1:l-1}, \bm{X}; \bm{\Theta}_{\tt mt}) ,
  \end{split}
\end{equation}
where $\bm{S} = \{s_1,\cdots,s_{|\bm{S}|}\}$ is the serialized token sequence and $s_l \in \{{\cal V} \cup {\cal O}\}$ is the $l$-th token in the serialized token sequence. $\cal O = \{{\tt [sep]}, {\tt [eos]}\}$ represents the special token set, where $\tt [sep]$ represents the speaker change and $\tt [eos]$ represents the end-of-sentence. Thus, we simply concatenate multiple textual tokens by inserting $\tt [sep]$ between utterances and insert $\tt [eos]$ at the end of the entire sequence. The serialized token sequence is represented as
\begin{multline}
  \bm{S} = \{w_1^1,\cdots,w_{N^1}^1, {\tt [sep]}, w_1^2,\cdots,w_{N^2}^2,  \\
  \cdots, w_{N^{T-1}}^{T-1},{\tt [sep]}, w_1^T,\cdots,w_{N^T}^T, {\tt [eos]}\} .
\end{multline}

In SOT for the multi-talker overlapped ASR with autoregressive modeling, the model parameter set can be optimized by
\begin{equation}
  \hat{\bm{\Theta}}_{\tt mt}  = \argmin_{\bm{\Theta}_{\tt mt}} - \sum_{(\bm{X},\bm{S}) \in {\cal D}_{\tt mt}}  \log  P(\bm{S}|\bm{X}; \bm{\Theta}_{\tt mt}) , 
\end{equation}
where ${\cal D}_{\tt mt}$ represents the multi-talker overlapped speech dataset. In this way, the autoregressive model is trained so as to recognize utterances of multiple speakers in the order of their start times, separated by a special symbol.

\section{Proposed Method}
We propose a unified autoregressive modeling for joint end-to-end multi-talker overlapped ASR and speaker attribute estimation. Our key idea is to handle gender and age estimation tasks within the unified autoregressive modeling. This enables us to effectively utilize the speaker attributes for improving multi-talker overlapped ASR.

\subsection{Modeling and its optimization}
In our joint end-to-end multi-talker overlapped ASR and speaker attribute estimation modeling, we predict the joint generation probability of multiple textual token sequences $\bm{W}^{1:T}$ and an individual speaker's attribute information from monaural overlapped speech $\bm{X}$. For the speaker attribute information, we jointly predict gender labels $\bm{g}^{1:T} = \{g^1,\cdots,g^T\}$ and age labels $\bm{a}^{1:T} = \{a^1,\cdots,a^T\}$ that correspond to multiple textual token sequences, where $g^t \in {\cal G}$ and $a^t \in {\cal A}$ are the $t$-th speaker's gender label and age label, respectively. ${\cal G} $ is the gender label set and ${\cal A}$ is the age label set, respectively. Thus, we handle the age estimation as not a regression problem but a classification problem. The joint generation probability is computed from
\begin{multline}
  P(\bm{W}^{1:T} , \bm{g}^{1:T}, \bm{a}^{1:T} |\bm{X}; \bm{\Theta}) \\
    \begin{split}      
      = & \prod_{t=1}^T P(\bm{W}^t , g^t, a^t |\bm{W}^{1:t-1}, g^{1:t-1}, a^{1:t-1}, \bm{X};  \bm{\Theta}) , \\
    \end{split}
\end{multline}
where $\bm{\Theta}$ represents the trainable model parameter sets of the joint modeling. In this way, a textual token sequence and speaker attribute labels for individual speakers are recursively generated one after another in an autoregressive manner. Thus, for estimating the $t$-th speaker's textual token sequence and speaker attributes, we can utilize not only all previous textual information but also all previous speaker attributes as contexts. The pre-estimated speaker attribute information should help us to estimate remaining utterances in the overlapped speech.

To efficiently model the joint generation probability using unified autoregressive modeling, we handle speaker and age labels as tokens as well as textual tokens. To this end, we serialize multiple textual token sequences and individual speaker's attributes into a single token sequence. Thus, we redefine the joint generation probability as
\begin{multline}
  P(\bm{W}^{1:T}, \bm{g}^{1:T}, \bm{a}^{1:T}  |\bm{X}; \bm{\Theta}) =  P(\bm{Z} |\bm{X}; \bm{\Theta}) \\
   =  \prod_{l=1}^{|\bm{Z}|} P(z_l|z_{1:l-1}, \bm{X}; \bm{\Theta}) ,
\end{multline}
where $\bm{Z}=\{z_1,\cdots,z_{|\bm{Z}|}\}$ is the serialized token sequence and $z_l \in \{{\cal V} \cup {\cal G} \cup {\cal A} \cup {\cal O}\}$ is the $l$-th token in the serialized token sequence. In fact, our main motivation is to improve multi-talker overlapped ASR performance using the speaker attribute estimation, so speaker attribute information should be utilized for estimating the textual information of not only future utterances but also current utterance. Therefore, we represent the serialized token sequence as
\begin{multline}
  \bm{Z} = \{g^1, a^1, w_1^1,\cdots,w_{N^1}^1, {\tt [sep]}, g^2, a^2, w_1^2,\cdots,w_{N^2}^2, \\
  \cdots,w_{N^{T-1}}^{T-1},{\tt [sep]}, g^T, a^T, w_1^T,\cdots,w_{N^T}^T, {\tt [eos]}\} .
\end{multline}
In this way, individual speaker attribute tokens are estimated before the corresponding textual information. In SOT for the joint end-to-end multi-talker overlapped ASR and speaker attribute estimation modeling, the model parameter set can be optimized by
\begin{equation}
  \hat{\bm{\Theta}}  = \argmin_{\bm{\Theta}} - \sum_{(\bm{X},\bm{Z}) \in {\cal D}} \log P(\bm{Z}|\bm{X}; \bm{\Theta}) ,
\end{equation}
where ${\cal D}$ represents the multi-talker overlapped speech dataset with speaker attributes.

\begin{figure}[t]
  \begin{center}
    \includegraphics[width=80mm]{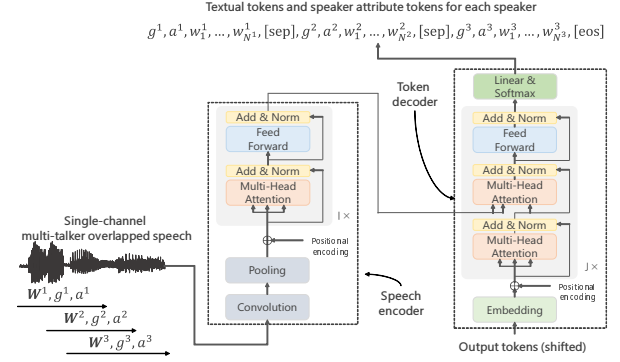}
  \end{center}
  \vspace{-6mm}
  \caption{Transformer-based implementation of our proposed method.}
  \vspace{-6mm}
\end{figure}

\subsection{Transformer-based implementation}
For the autoregressive modeling, this paper uses a transformer \cite{vaswani_nips2017}. In our transformer-based autoregressive modeling, we compute $P(z_l|z_{1:l-1}$, $\bm{X}$; $\bm{\Theta})$ using a speech encoder and a token decoder. Figure 1 shows the transformer-based implementation of joint end-to-end multi-talker overlapped ASR and speaker attribute estimation.

Our speech encoder converts input acoustic features $\bm{X}$ into the hidden representations. To this end, convolution layers and pooling layers are first introduced for producing subsampled speech representations from the input acoustic features. Next, position information is embedded into the subsampled speech representations. After that, multiple transformer encoder blocks are used for taking surrounding contexts into consideration. Note that one transformer encoder block consists of a scaled dot product multi-head self-attention layer and a position-wise feed-forward network \cite{vaswani_nips2017}. Our token decoder computes $P(z_l|z_{1:l-1}, \bm{X}; \bm{\Theta})$, i.e., the generative probability of a token given preceding tokens $z_{1:l-1}$ and the hidden representations produced in the speech encoder. First, the preceding tokens are embedded into continuous representations using a linear embedding layer. Next, position information is embedded into the continuous representations of the tokens. After that, multiple transformer decoder blocks are used for taking both the preceding token contexts and input speech contexts into consideration. Note that one transformer decoder block consists of a scaled dot product multi-head masked self-attention layer, a scaled dot product multi-head source-target attention layer, and a position-wise feed-forward network \cite{vaswani_nips2017}. Lastly, predicted probabilities for the $l$-th token $z_l$ are calculated in a softmax layer with a linear transformation.

\setcounter{table}{1}
\begin{table*}[t!]
  \begin{center}
      \caption{Experimental results of multi-talker overlapped ASR and speaker attribute estimaion.}
      \footnotesize
      \vspace{-3mm} 
      \begin{tabular}{|c|l|rrrrr|rrr|} \hline
        \# of speakers & Multi-talker overlapped ASR systems &  \multicolumn{5}{|c|}{CER [\%]} & SCA [\%] & SGA [\%] & SAA [\%] \\ 
        in test dataset & & 1 & 2 & 3 & 1+2 & 1+2+3 & 1+2+3 & 1+2+3 & 1+2+3 \\ \hline \hline
        1 & Conventional & 5.23 & 9.70 & 15.23 & 4.98 & 4.75 & {\bf 99.99} & - & - \\
        & Proposed (Gender) & 4.99 & 8.84 & 14.11 & 4.72 & 4.65 & {\bf 99.99} & 99.53 & - \\ 
        & Proposed (Age) & 5.18 & 9.52 & 15.08 & 4.92 & 4.74 & {\bf 99.99} & - & 53.11 \\
        & Proposed (Gender \& Age) & 4.92 & 8.70 & 14.05 & 4.75 & {\bf 4.62} & {\bf 99.99} & {\bf 99.69} & {\bf 53.15} \\ \hline \hline
        2 & Conventional & 38.03 & 9.49 & 11.66 & 9.56 & 8.24 &  98.54 & - & - \\
        & Proposed (Gender) & 37.45 & 9.37 & 11.46 & 9.35 & 7.90 & 99.10 & {\bf 98.22} & - \\
        & Proposed (Age) & 37.80 & 9.45 & 11.58 & 9.52 & 8.12 & 98.62 & - & 53.47 \\
        & Proposed (Gender \& Age) & 37.04 & 9.25 & 11.42 & 9.18 & {\bf 7.68} & {\bf 99.19} & {\bf 98.82}& {\bf 53.67} \\ \hline \hline
        3 & Conventional & 51.92 & 34.49 & 15.41 & 34.37 & 16.09 & 94.07 & - & - \\ 
        & Proposed (Gender) & 51.43 & 33.21 & 14.23 & 33.75 & 15.05 & 94.53 & 96.40 & - \\ 
        & Proposed (Age) & 51.70 & 34.06 & 15.10 & 34.10 & 15.75 & 94.23 & - & 51.93 \\
        & Proposed (Gender \& Age) & 51.20 & 32.91 & {\bf 13.95} & 33.64 & 14.68 & {\bf 95.04} & {\bf 97.32} & {\bf 52.30} \\ \hline 
      \end{tabular}
  \end{center}
  \vspace{-6mm}
\end{table*}

\section{Experiments}
In experiments, we used the Corpus of Spontaneous Japanese (CSJ) \cite{maekawa_lrec2000}, which is a dataset for single-talker ASR. We divided the CSJ into training, validation, and test datasets. Note that each lecture audio-signal was segmented into utterance-level audio signals. We call them single-speaker datasets. Table 1 show the detailed information. For examining multi-talker overlapped ASR, we mixed multiple audio signals as a monaural signal. To this end, we randomly chose multiple audio signals from each dataset so as not to select the same speakers. We set the number of speakers in the mixed signals as two or three. When mixing the audio signals, the original volume of each utterance was kept unchanged, resulting in an average signal-to-interference ratio of about 0 dB. As for the delay applied to each utterance, the delay values were randomly chosen under the constraints the same as previous work \cite{kanda_interspeech2020a}. First, the start times of the individual utterances differ by 0.5 s or longer. Second, every utterance in each mixed audio sample has at least one speaker-overlapped region with other utterances. In fact, the number of audio signals in two-speaker or three-speaker datasets is same as that in the original single-speaker datasets. For ASR, this paper used characters as the textual tokens.

\subsection{Setups}
We constructed various single-talker ASR systems and multi-talker overlapped ASR systems. Conventional methods are systems that did not utilize speaker attributes as described in Section 3. Proposed methods are systems that utilize single- or multiple-speaker attribute estimation tasks, i.e., gender and age estimation. By using single-speaker, two-speaker, and three-speaker training datasets, we constructed several systems. ``1'' represents single-talker ASR systems trained from single-speaker datasets. ``1+2+3'' represents multi-talker overlapped ASR systems trained from the mixture of 1, 2, and 3 speakers. For modeling speaker attributes, we defined gender labels as either male or female, and age labels as 20-class generation labels between 0 and 100 years old.

\setcounter{table}{0}
\begin{table}[t!]
 \label{}
 \begin{center}
   \caption{Detailed information of single-speaker datasets.}
   \vspace{-3mm}
   \footnotesize
   \begin{tabular}{|l|rrrr|} \hline
    & Data size & \# of & \# of  & \# of \\
    & (Hours)  & signals & characters & speakers   \\ \hline \hline
    Train & 518.4 & 417,406 & 13,471,877 & 1,430 \\
    Vald & 1.3 & 1,385 & 32,089 & 10 \\
    Test & 1.9 & 1,292 & 47,970 & 10 \\  \hline
  \end{tabular}
 \end{center}
 \vspace{-6mm}
\end{table}

For the transformer-based autoregressive models, the transformer blocks were composed under the following conditions: the dimensions of the output continuous representations were set to 512, the dimensions of the inner outputs in the position-wise feed-forward networks were set to 2,048, and the number of heads in the multi-head attentions was set to 4. In the nonlinear transformational functions, the Swish activation was used. For the speech encoder, we used 40 log mel-scale filterbank coefficients appended with delta and acceleration coefficients as acoustic features. The frame shift was 10 ms. The acoustic features passed two convolution and max pooling layers with a stride of 2, so we down-sampled them to $1/4$ along with the time axis. After these layers, we stacked 4-layer transformer encoder blocks. In the text decoder, we used 512-dimensional character embeddings where the vocabulary size was set to 3,262. We also stacked 3-layer transformer decoder blocks. For the training, we used the RAdam optimizer \cite{liu_iclr2020}. The training steps were stopped on the basis of the early stopping using part of the training data. We set the mini-batch size to 64 utterances and the dropout rate in the transformer blocks to 0.1. We introduced label smoothing where its smoothing parameter was set to 0.1. In addition, we applied SpecAugment \cite{park_is2019}. Our SpecAugment only applied frequency masking and time masking. For testing, we used a beam search algorithm in which the beam size was set to 4.

\subsection{Results}
Table 2 shows the results in terms of character error rate (CER) for each ASR system and speaker counting accuracy (SCA), speaker gender accuracy (SGA), speaker age accuracy (SAA) for multi-talker overlapped ASR systems trained by the mixture of 1, 2, or 3 speakers (``1+2+3''). In multi-talker overlapped ASR setups, we compared references with hypotheses while considering the order of utterances. Note that we only evaluated textual tokens except for special tokens and speaker attribute tokens to compute CER.

First, the results show that multi-talker overlapped ASR systems (``2'', ``3'', ``1+2'', and ``1+2+3'') well recognize multi-talker overlapped speech of known number of speakers while single-talker ASR systems (``1'') cannot handle them at all. Next, the results show that proposed methods that jointly modeled with speaker attribute estimation outperformed conventional methods in each test setup. Especially, the proposed methods were effective when the number of speakers in the test dataset was large. This indicates that speaker attributes were effectively utilized for improving multi-talker overlapped ASR problems. In addition, the proposed method with gender estimation outperformed that with age estimation in each test setup. This is considered to be because gender information is more effective context information to find multiple utterances in multi-talker overlapped speech. The best results were attained by the proposed method jointly utilizing both gender and age estimation. This indicates that consideration of multiple-speaker attributes is effective for solving multi-talker overlapped ASR problems. Furthermore, SCA, SGA and SAA were improved by considering multiple attributes when the number of speakers in test dataset was large. Thus, consideration of multiple-speaker attributes has been shown to be beneficial for being robust against speaker confusion. These results demonstrate that our proposed method effectively improves multi-talker overlapped ASR performance.

\section{Conclusions}
We presented unified autoregressive modeling for joint end-to-end multi-talker overlapped ASR and speaker attribute estimation. The key strength of the proposed method is that we can utilize not only the textual information but also speaker attributes as contexts in a single-channel multi-talker overlapped ASR system. This is implemented by handling gender and age estimation tasks within the unified autoregressive modeling. Our experimental results showed that jointly estimating gender and age information for each utterance improves multi-talker overlapped ASR performance.

\end{document}